# DIVIDE: A Framework for Learning from Independent Multi-Mechanism Data Using Deep Encoders and Gaussian Processes

Vivek Chawla[1*], Boris Slautin[2], Utkarsh Pratiush[2], Dayakar Penumadu[1], Sergei Kalinin[2]

[1] Department of Civil and Environmental Engineering, University of Tennessee, Knoxville, TN 37996, USA
[2] Department of Materials Science and Engineering, University of Tennessee, Knoxville, TN 37996, USA

## ABSTRACT

Scientific datasets often arise from multiple independent mechanisms such as spatial, categorical or structural effects, whose combined influence obscures their individual contributions. We introduce DIVIDE, a framework that disentangles these influences by integrating mechanism-specific deep encoders with a structured Gaussian Process in a joint latent space. Disentanglement here refers to separating independently acting generative factors. The encoders isolate distinct mechanisms while the Gaussian Process captures their combined effect with calibrated uncertainty. The architecture supports structured priors, enabling interpretable and mechanism-aware prediction as well as efficient active learning. DIVIDE is demonstrated on synthetic datasets combining categorical image patches with nonlinear spatial fields, on FerroSIM spin lattice simulations of ferroelectric patterns, and on experimental PFM hysteresis loops from $PbTiO_3$ films. Across benchmarks, DIVIDE separates mechanisms, reproduces additive and scaled interactions, and remains robust under noise. The framework extends naturally to multifunctional datasets where mechanical, electromagnetic or optical responses coexist.

## INTRODUCTION

Many real-world systems exhibit behavior driven by the combined influence of multiple independent mechanisms. These mechanisms may represent categorical factors, spatial dependencies, or nonlinear physical responses. While the scalar output of such systems is observable, the individual contributions of these mechanisms are often unknown and unmeasured. Modeling this type of data requires not only accurate predictions but also the ability to attribute variation in the output to specific, distinct sources. In this context, we use disentanglement to mean recovering those independently acting generative factors from observational data. Disentangling these contributions is particularly important in scientific and engineering domains where interpretability, causality, and mechanism-aware reasoning are essential.

Partial solutions to this challenge have emerged from the field of disentangled representation learning, which seeks to identify independent factors of variation from high-dimensional data. Approaches such as β-VAE[1], FactorVAE[2], and InfoGAN[3] introduce regularization techniques to encourage statistical independence in the latent space. However, while effective in unsupervised scenarios, these methods often rely on assumptions, such as minimal total correlation or alignment with factorized priors, that do not hold in real-world settings. Those familiar with the causality literature[4] will recognize that there are infinitely many generative models that can yield the same marginal distribution over observations, $p(x)$, but differ in their underlying mechanisms or causal structure. Without additional constraints or architectural separation, the model has no incentive to learn a disentangled or identifiable representation.

* Corresponding author. Tel: +1 315-800-8228 E-mail address: vchawla@vols.utk.edu

Empirical studies have shown that minimizing KL divergence or total correlation alone does not always lead to semantically meaningful disentanglement[5]. In contrast, many scientific and engineering domains offer partial knowledge about the underlying mechanisms. For instance, when it is known that two generative factors: such as spatial location and material phase operate independently, this structural prior can serve as a valuable inductive bias. Recent works have demonstrated that incorporating such domain knowledge into model architecture improves both generalization and interpretability, particularly in supervised learning settings[6-9].

Parallel developments in variational inference have embraced similar ideas. Structured Mean-Field Variational Inference (SMFVI)[10] relaxes the fully factorized assumption of traditional mean-field approaches by allowing sample-wise correlations within each latent dimension, while maintaining independence across dimensions. This structure is especially effective for grouped observations, where dependencies are localized to specific generative factors. Similarly, multimodal VAEs[11] and other modular architectures enforce architectural disentanglement by assigning distinct encoders to different modalities or mechanisms, preserving separation during both training and inference.

Gaussian Processes (GPs) are especially well-suited for supervised tasks due to their flexibility and calibrated uncertainty estimates. To address this, methods such as Deep Gaussian Processes with Decoupled Inducing Inputs[12] and Additive GPs[13] introduce architectural and kernel-based mechanisms to disentangle distinct influences, particularly effective when inputs are low-dimensional or mechanisms are known a priori. In automated experimentation, Deep Kernel Learning (DKL) continues to gain popularity for its ability to scale GPs to high-dimensional and multimodal data[14], and for generating target-dependent latent embeddings-in contrast to VAEs, which typically produce static, input-dependent representations. By combining the expressiveness of deep networks with the principled uncertainty estimation of GPs, it enables efficient, calibrated learning even from limited labeled data. DKL has been applied in automated experimentation[15,16], structure-property modeling[17], atomic force microscopy[18], scanning probe microscopy[8,9,19-21], and spectroscopic imaging[22], where it helps construct low-dimensional latent manifolds that guide intelligent data acquisition. However, existing DKL models typically rely on a single encoder and do not explicitly separate different sources of variation, risking the entanglement of unrelated generative factors.

For automated scientific discovery, solving such disentanglement problems is paramount. Many experimental systems produce scalar outputs that are influenced by multiple independent physical mechanisms. For instance, understanding residual stress variation remains one of the most significant challenges in additive manufacturing, where thermal gradients and solidification processes introduce complex, spatially varying stress fields[23]. Nanoindentation can be used to probe mechanical properties in such contexts, but the measured hardness can be simultaneously affected by residual stress and local crystallographic anisotropy. Disentangling these contributions is essential for accurate interpretation [24,25]. Similarly, in ferroelectric materials, scanning probe techniques such as piezo response force microscopy (PFM) must separate the electromechanical response from topography or local conductivity variations to correctly resolve domain structure and switching behavior[26,27]. Structured models that explicitly reflect these independent mechanisms offer a path toward more interpretable and mechanism-aware learning.

To address this gap, we propose DIVIDE: Disentangling Influences via Variational Inference and Deep Encoding. DIVIDE is a modular framework that combines mechanism-specific encoders with a structured Gaussian Process (GP) defined over their concatenated latent representations. Each encoder is responsible for capturing a distinct generative factor, preserving independence by design. This leads to interpretable latent spaces, accommodates structured priors, and supports uncertainty-aware prediction. By aligning model structure with known independence in the data, DIVIDE improves both representational quality and predictive performance.

We evaluate DIVIDE on two synthetic benchmarks constructed to simulate realistic disentanglement scenarios. These include RGB color patches (representing discrete categorical mechanisms), suit-based image patches[15] (introducing transformations and structural variations), and spatial fields composed of overlapping tensile and compressive Gaussian peaks (modeling continuous variation). By combining individual generative factors, we test the model's ability to recover and isolate each contributing influence. Our results show that DIVIDE performs robustly across a wide range of settings. In single-mechanism datasets, it captures both categorical and spatial dependencies with high fidelity. In multi-mechanism cases, it preserves separation in latent space and accurately attributes variation in the output. When structured priors are introduced, for complex spatial fields, the model converges faster and produces smoother, more accurate predictions. Under active learning, DIVIDE consistently selects informative inputs, improving both data efficiency and predictive stability compared to monolithic models.

Additionally, we investigated the local ferroelectric properties as a further benchmark of the DIVIDE workflow. The process of spontaneous polarization switching in ferroelectrics is governed by a wide range of sample characteristics, including the domain arrangement. Consequently, optimizing domain structures to improve the target functionality is an important task for numerous applications, such as micromechanical systems[28], sensors[29], energy-harvesting devices,[30] and beyond. The $PbTiO_3$ (PTO) film was used as a model system. Its microstructure comprises both in-plane ferroelastic $a$-domains and out-of-plane ferroelectric $c$-domains, offering a high degree of variability in domain configurations. The nontrivial domain structure often complicates optimization, as it hidden the individual contributions of different domain types and their associated properties to the switching behavior. In this context, the DIVIDE approach is particularly promising, as it enables the deconvolution of these influences, allowing a clearer understanding of the individual impact of different domain types and properties and facilitating domain engineering. The evaluation of the DIVIDE framework for ferroelectric properties was carried out using two case studies of different complexity: first, we assessed its efficacy with the artificial FerroSIM[31,32] model for 2D ferroelectrics, and subsequently with real PFM data measured on PTO thin films.

In summary, DIVIDE provides a principled and modular approach for disentangling independent data-generating mechanisms from observational data. By combining deep, mechanism-specific encoders with a structured GP regression model, it supports interpretable, uncertainty-aware learning. The framework is broadly applicable to scientific problems where outputs reflect interacting but separable sources, and where understanding those sources is as important as making accurate predictions. This modularity also makes DIVIDE applicable to multifunctional materials where mechanical, electrical, magnetic, or optical properties arise from overlapping but independent mechanisms

**RESULTS AND DISCUSSION**
**Background**
A Gaussian Process (GP) treats an unknown function as a single, interconnected random surface: any set of inputs is assumed to share a joint Gaussian distribution whose kernel encodes how strongly one point "talks" to another (e.g., the Matern kernel for rough signals). Conditioned on data, a GP yields both a prediction and an uncertainty band. However, the computational cost escalates sharply as the dimensionality grows. Deep Kernel Learning (DKL) [33] tackles this by first pushing inputs through a neural-network encoder to learn task-specific features, then applying a GP in that latent space. Training uses a variational Evidence Lower Bound (ELBO) with inducing points, letting the model combine deep feature learning with GP-style calibrated uncertainty on large datasets.

Despite this, a single, monolithic encoders often entangle latent factors, making it difficult to trace output variation back to independent mechanisms. Our architecture avoids this by design, ensuring that predictive uncertainty and interpretability are preserved through architectural separation and constraint enforcement. Assuming each encoder $f_k(x; \theta_k)$ is continuous in both its input and parameters, the resulting latent mapping $z(x; \theta)$ is also continuous in $\theta$. Because both the GP kernel $K(z, z')$ and the mean function $m(z)$ are continuous by design, the GP's predictive mean and variance also converge pointwise as training progresses:

$$z(x; \theta) \rightarrow z(x; \theta^*) \text{ for all } x, \tag{1}$$
$$K(z(x; \theta), z(x'; \theta)) \rightarrow K(z(x; \theta^*), z(x'; \theta^*)), \tag{2}$$
$$m(z(x; \theta)) \rightarrow m(z(x; \theta^*)) \tag{3}$$

This behavior ensures that, near a well-tuned set of weights $\theta^*$, a small tweak $\Delta\theta$ only produces a small change in the output $f(x; \theta)$. But because the loss surface of a neural network is non-convex, training isn't guaranteed to reach that well-tuned region in the first place. A critical but often implicit requirement for disentanglement is injectivity (i.e., one-to-one mapping) in the encoder mapping. If the encoder maps distinct input configurations to the same latent code despite differing outputs, the GP is exposed to conflicting supervision at that point in latent space, leading to poor predictions and high uncertainty[34]. By separating mechanisms architecturally, the model better preserves identifiability and encourages interpretable representation learning.

Although the design promotes disentanglement, convergence and injectivity are not guaranteed. In principle, additional regularization techniques could further improve separation. In this work, however, we rely solely on encoder separation and normalization to achieve mechanism-specific encoding. The GP kernel is typically invariant to global translations in the latent space. Without additional constraints, this introduces non-uniqueness in the decomposition. For example, shifts of the form $z_k(x) \mapsto z_k(x) + \delta_k$, where $\sum_{k=1}^{n} \delta_k = 0$, leave the GP unchanged. To resolve this ambiguity, we normalize each latent and impose $n - 1$ anchor constraints. These constraints make the decomposition unique up to rotation and allow interpretable attribution of output variation to individual mechanisms. A schematic of the proposed architecture, including the mechanism-specific encoders and GP regression over the joint latent space, is shown in Figure 1. In addition to these high-dimensional inputs (e.g., image patches), we often encounter auxiliary low-dimensional data that carry physical meaning-such as the $(x, y)$ spatial location of each patch, or other contextual variables like local composition or boundary proximity. When such information is available and meaningful, it can be directly integrated into the latent space before applying the Gaussian process (Figure 1). The complete procedure is outlined in Pseudocode 1 (Table 1).

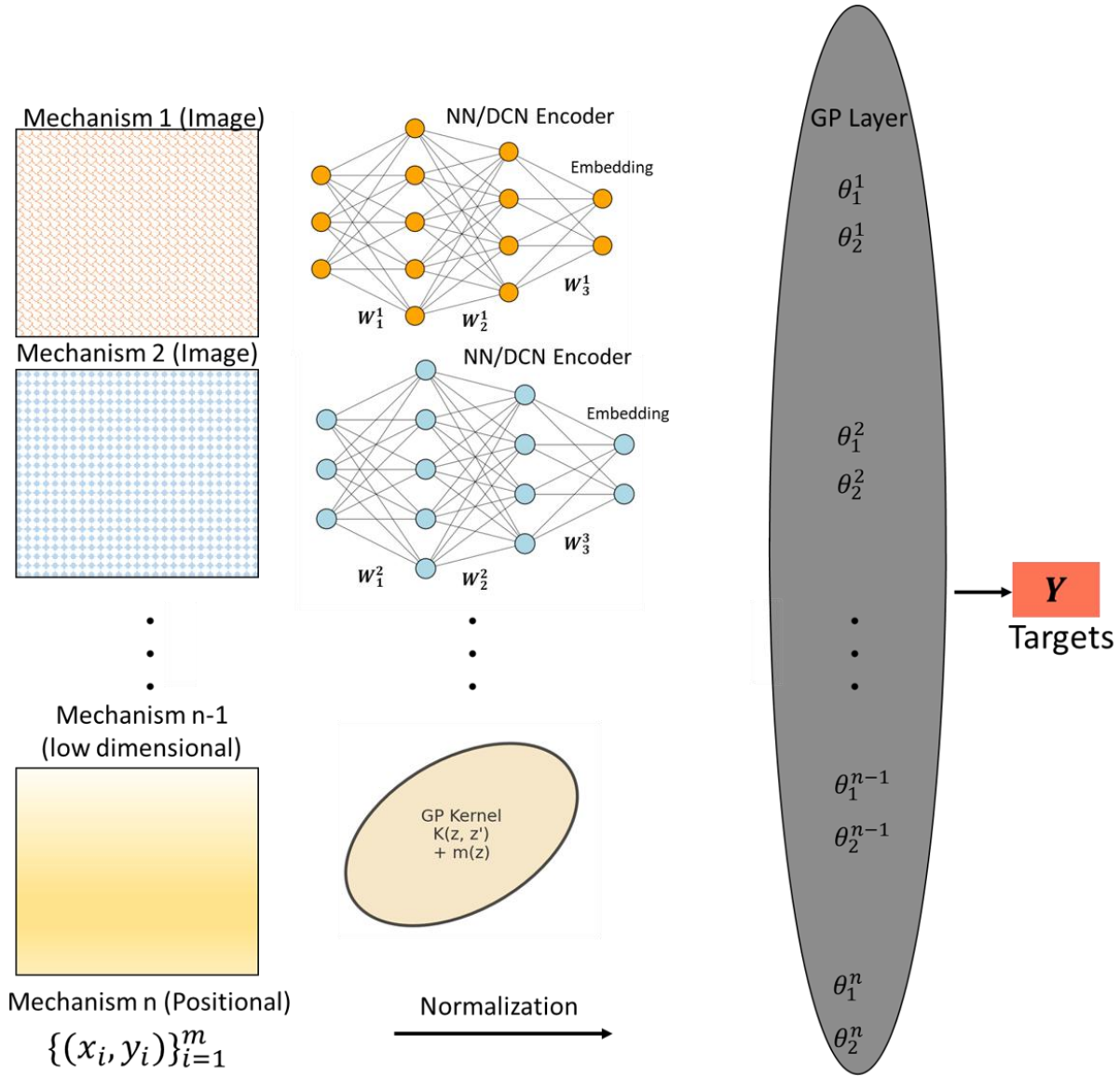

**Figure 1:** Overview of the DIVIDE architecture. Each independent mechanism is processed by a dedicated encoder (e.g., CNN or ANN), producing mechanism-specific latent features. These are concatenated to form a joint latent vector. Low-dimensional contextual variables, such as normalized spatial coordinates (x, y), are appended directly to the latent space. The combined representation is passed to a Gaussian Process that models the scalar output.

**Table 1:** Pseudocode 1. DIVIDE setup

| Step | Operation | Description |
| --- | --- | --- |
| 1 | Acquire | Obtain multi-modal image inputs $\{I_1, I_2, \ldots, I_m\}$ corresponding to distinct mechanisms. |

| 2 | Segment | Extract image patches $X_j$ from each modality $I_j$ using patch size u. |
| 3 | Normalize | Normalize physical spatial coordinates $(x_i, y_i)$ to lie in the range $[-1, 1]^2$ for each location in the grid. |
| 4 | Encode | Pass image patches through mechanism-specific CNNs to extract latent features. Concatenate all latent outputs and directly append normalized spatial coordinates $(x_i, y_i)$ within the latent space to form the final input vector. |
| 5 | Prior (Optional) | If available, define a structured prior mean over $(x, y)$ to be used with the GP for spatial modeling (e.g., boundary constraints). |
| 6 | Initialize | Select initial seed set $S_0$ of size $k_0$ using either random or policy-driven sampling. |
| 7 | Measure | Measure scalar outputs $y$ at seed locations: $y_{S0} = \text{measure}(S_0)$. |
| 8 | Train | Train a Gaussian Process (GP) or structured GP (sGP) on latent spatial + inputs. Use the structured prior mean if specified. |
| 9 | Predict | Predict posterior mean and uncertainty at all unsampled points: $\mu, \sigma = GP.predict\left(Input_{grid \setminus S_n}\right)$ |
| 10 | Score | Compute acquisition scores: $\alpha_i = \sigma_i - \lambda.penalty(x_i, y_i)$, where the penalty discourages selection near boundaries. |
| 11 | Select | Select the next sampling point as: $i^* = argmax\ \alpha_i$ |
| 12 | Measure | Measure output at $i^*$, update: $S_{n+1} = S_n \cup S_{i^*}$ & $y_{n+1} = y_n \cup y_{i^*}$ |
| 13 | Iterate | Repeat steps until budget (100 iterations) is exhausted or convergence is reached. |

**Benchmark 1: Accuracy and Mechanism Disentanglement**

To model the first benchmark, we extract a subset of image patches along with their spatial coordinates $(x, y)$, normalize the coordinates, and pass both inputs through a hybrid architecture. The image patch is encoded via a convolutional neural network (CNN) into a low-dimensional latent representation, while the normalized spatial coordinates are directly appended to this latent vector and passed to a Gaussian Process (GP) for regression. The full model architecture is provided in the Supporting Information (SI Section S2). The image patches are encoded via a convolutional neural network, while the spatial coordinates are appended directly to the latent representation before being passed to a Gaussian process (GP). The model is trained using a variational ELBO objective with inducing points drawn from a subset of the training data.

As shown in Figure 2a, the target for the model is derived using the summation of two independent mechanisms. The model learns a smooth predictive mean that reflects the additive nature of the true target field (Figure 2b, compare ground truth and predictions). To evaluate whether the model disentangles the two mechanisms, we conduct two controlled tests. In the first, we fix the spatial coordinates $(x, y)$ and allow the image patch to vary. This isolates the effect of the visual mechanism. When the GP predictions are clustered into three groups using KMeans, the resulting clusters correspond closely to the three original color classes, indicating that the model has learned a patch-dependent latent representation (Figure 2b). This serves as a good sanity check, confirming that the model captures the intended visual distinctions. In the second test, we fix the image patch and allow only the spatial coordinates to vary. The resulting predictions reveal a smooth gradient that closely matches the original linear spatial trend, confirming that the spatial mechanism has been captured independently of image variation (Figure 2b).

While the above tests demonstrate empirical evidence of mechanism disentanglement, the decomposition remains underdetermined without additional constraints. As previously established, disentangling $n$ independent mechanisms require $n-1$ independent constraints to ensure identifiability. In the current benchmark, where the scalar output is formed from the sum of two mechanisms (one spatial and one patch-based), a single constraint is required to uniquely separate their contributions. To enforce this constraint, we apply a fixed anchor point at a known reference location. We define this anchor by assuming that it is known that the spatial mechanism contributes zero at the coordinate $(0,0)$. In many physical systems, knowing the value at one location is typically not an issue, as solutions are often prescribed or well-defined at boundaries or in the far-field regions. Importantly, the total target value $y(0,0)$ at this point is not assumed to be zero; rather, it is entirely attributed to the local image-based mechanism. By subtracting the predicted output at this anchor from the full GP solution, we eliminate the arbitrary spatial offset and reassign it to the patch-based component. This adjustment results in a disentangled representation in which the spatial trend begins at zero and both mechanisms retain their relative variations. The fixed anchor point thereby enforces a valid decomposition under the identifiability condition and ensures a consistent interpretation of each mechanism's contribution. As a result, the recovered solution aligns precisely with the true ground truth decomposition, confirming that the model has learned the correct latent structure. The full procedure for evaluating mechanism disentanglement, by selectively varying one input while holding others fixed and applying anchor-based corrections, is detailed in Pseudocode 2 (Table 3).

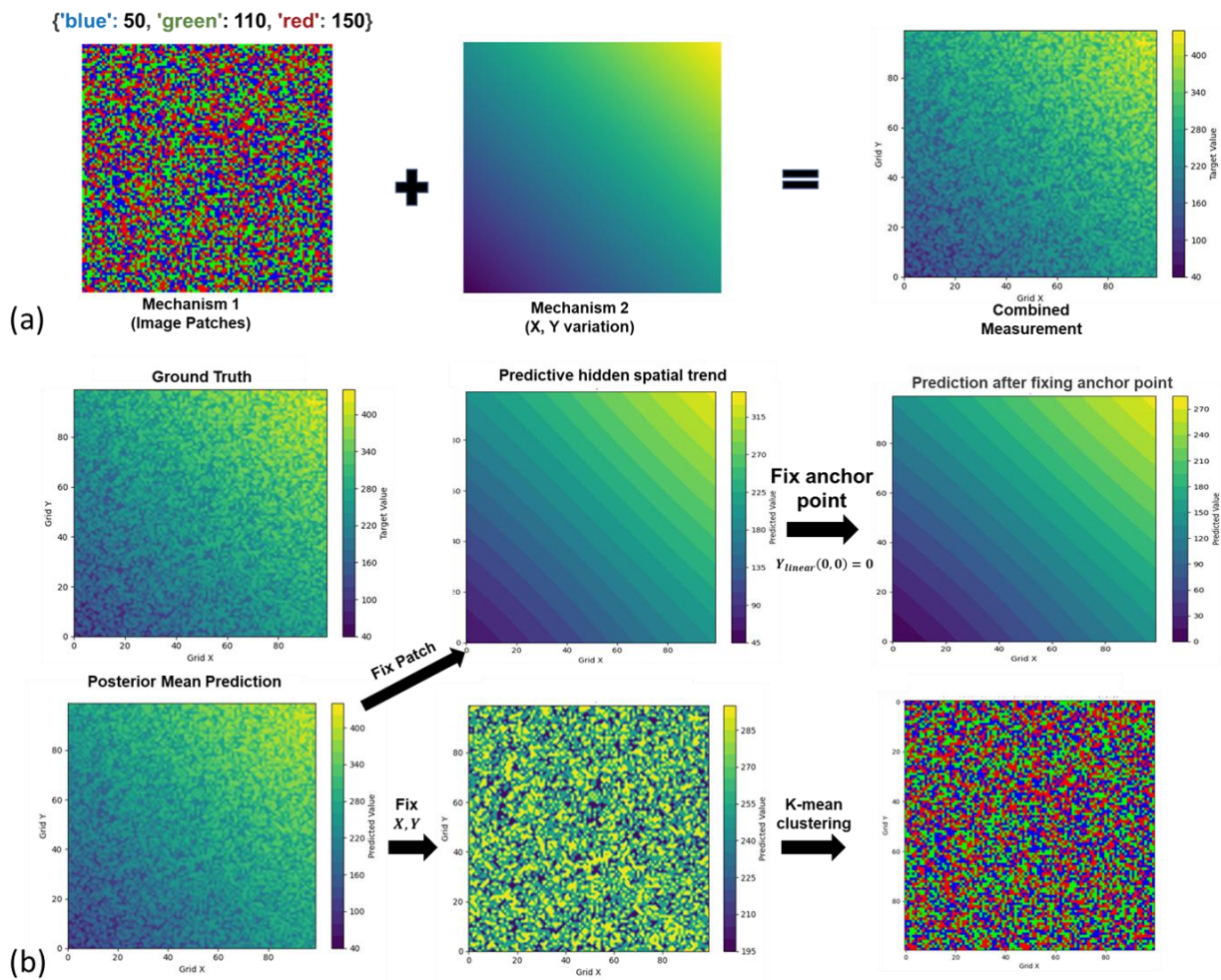

**Figure 2:** Disentangling spatial and visual mechanisms. The GP learns a smooth additive field from image patches and spatial coordinates. Controlled tests confirm separation: (a) fixed spatial input yields patch-dependent predictions matching color classes; (b) fixed patch input recovers the spatial trend. An anchor constraint at (0,0) ensures identifiability, enabling accurate decomposition of both mechanisms.

Table 2: Pseudocode 2. Disentanglement Evaluation via Controlled Mechanism Inputs

| Step | Action | Description |
|---|---|---|
| 1 | Select active mechanism | Choose one mechanism (image-based or structured) to isolate. |
| 2 | Fix other mechanisms | For all other mechanisms:<br>- If image-based: fix input to a single representative patch.<br>- If spatial (x, y): fix coordinates to a constant |
| 3 | Evaluate model | Pass full dataset with only the active mechanism varying. Record predicted output over the grid. |
| 4 | Apply anchor correction | Fix one input condition (e.g., known value for x,y at (0,0)), and subtract predicted value at this location to align outputs. This resolves additive ambiguity. |

| 5 | Repeat for all mechanisms | Repeat steps 1–4 for each mechanism independently. |
| 6 | Final mechanism constraint | For the last mechanism (e.g., spatial), apply the constraint: $\sum_k \delta_k = 0$, to compute the residual shift $\delta_k$ and align the final output. |
| 7 | Compare to ground truth | Compare isolated predictions to known component fields (if available) to evaluate disentanglement accuracy. |

Given that the model was able to disentangle the two mechanisms. Furthermore, we applied pure uncertainty based active learning from the GP over spatial and patch-derived features. A spatial penalty was used to avoid edge-biased sampling (SI section S3), encouraging informative points from the interior. Initially, the model captures some global variation through $x, y$ and some local variation through patches, but the resulting prediction is far from the true behavior. As active learning progresses, the model begins to favor a globally smooth explanation dominated by $x, y$, effectively ignoring the patch-based contribution. However, by around iteration 10, the model starts to reconcile both global and local effects, significantly reducing uncertainty and closely approaching the true underlying behavior. (Figure 3; see SI Section S3 for details). Moreover, we evaluated the robustness of active learning with DIVIDE using two scenarios: no spatial contribution (no x and y effect), and in case with high noise wherein both scenarios, algorithm converged (see SI section S4 for details).

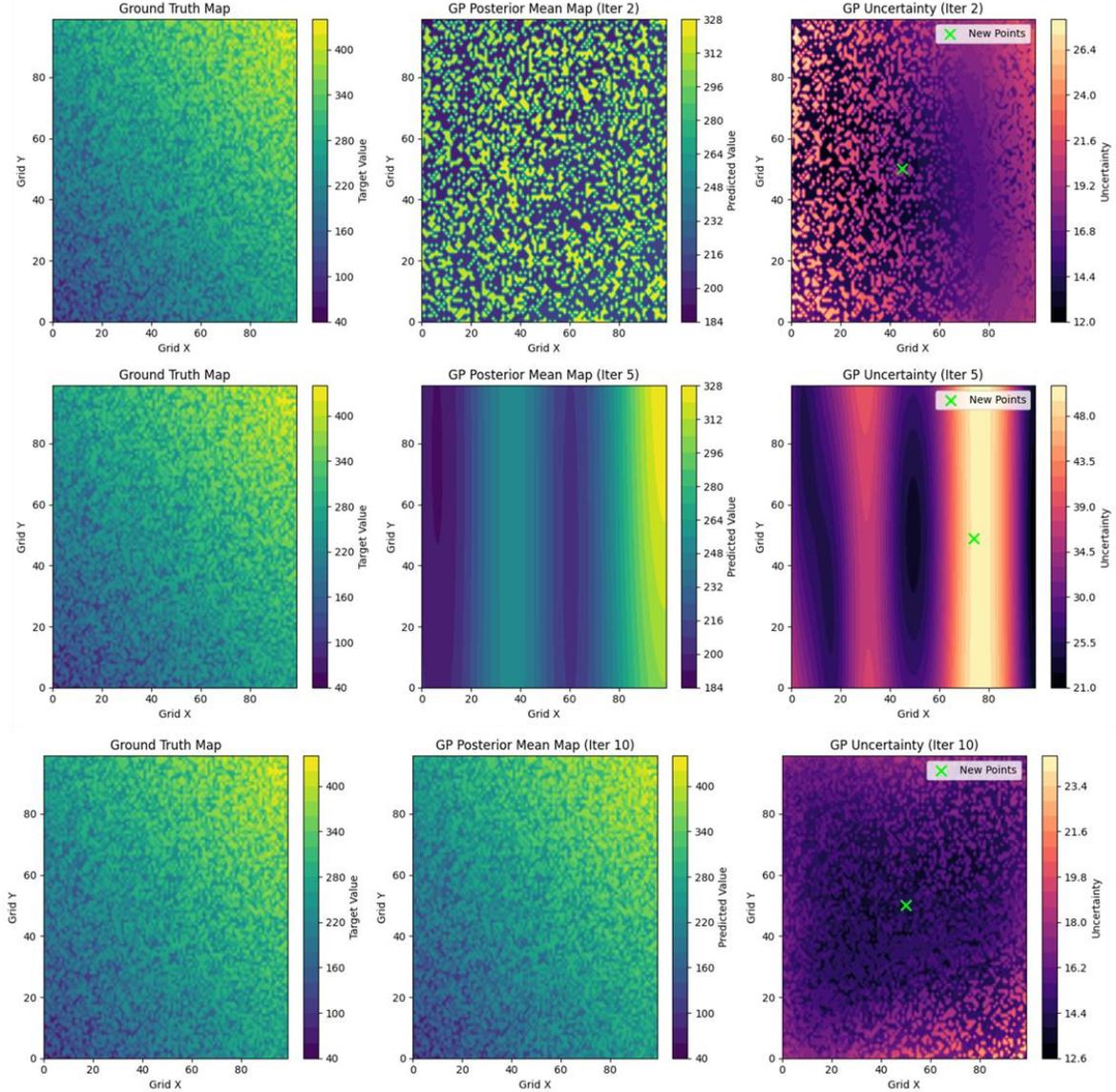

**Figure 3:** Active learning with two initial points. Early predictions are poor, but by iteration 10, the model captures both spatial and patch-based effects. A spatial penalty guides sampling away from boundaries, improving learning efficiency.

**Benchmark 2: Structured GP for Nonlinear Spatial Fields**

Before moving to the next case involving more complex spatial field and image patches, it was established that the standard GP fails in the presence of the complex residual stress spatial field (see SI section 5). Therefore, to address the failure of standard GPs in capturing the complex spatial variation, we utilize Structured Gaussian Process (sGP) that incorporates domain knowledge directly into the model through a custom prior mean function. Based on physical intuition from residual stress behavior in welded materials, we assume the presence of a Gaussian-like tensile peak in the spatial domain. This is encoded by defining a structured mean function of the form:

$$\mu(x, y) = A * \exp\left(-\frac{(x - x_c)^2 + (y - y_c)^2}{2\omega^2}\right) \tag{4}$$

where $(x_c, y_c)$ is the center of the peak, $A$ is the amplitude, and $\omega$ is the width (standard deviation). These parameters are treated as learnable and adapt during training. Importantly, this structured mean is applied only to the spatial components (i.e., the last two dimensions corresponding to $x$ and $y$ of the input), while the suit-based mechanism is modeled with a separate constant mean. Figure 4a shows the input mosaic of suit patches along with the underlying stress distribution. Figure 4b shows the progression of the model's solution across active learning iterations. Initially, predictions reflect the influence of the prior mean, which assumes a Gaussian peak centered in the domain. This is highlighted as an inner circular pattern in the predicted spatial variation. By iteration 25, the model has successfully learned to classify the suit-based image patches, but the spatial component remains largely unchanged from the initial prior. Only at iteration 33 does the model begin to correctly identify the center of the tensile peak. Model is able to recover the high-magnitude positive peak effectively, while the broader and lower-magnitude compressive region is less accurately represented.

Figure 4c shows the final clustering of image patches according to suit type. The corresponding recovered spatial field is also shown, demonstrating partial success in resolving both underlying mechanisms. This result underscores the utility of incorporating structured priors while also illustrating the remaining difficulty in fully capturing multi-scale spatial signals. Finally, a case with three mechanisms (RGB mosaic, Suit mosaic, and Gaussian spatial variation) was also tested with active learning that also converged (see SI section S6 for more details).

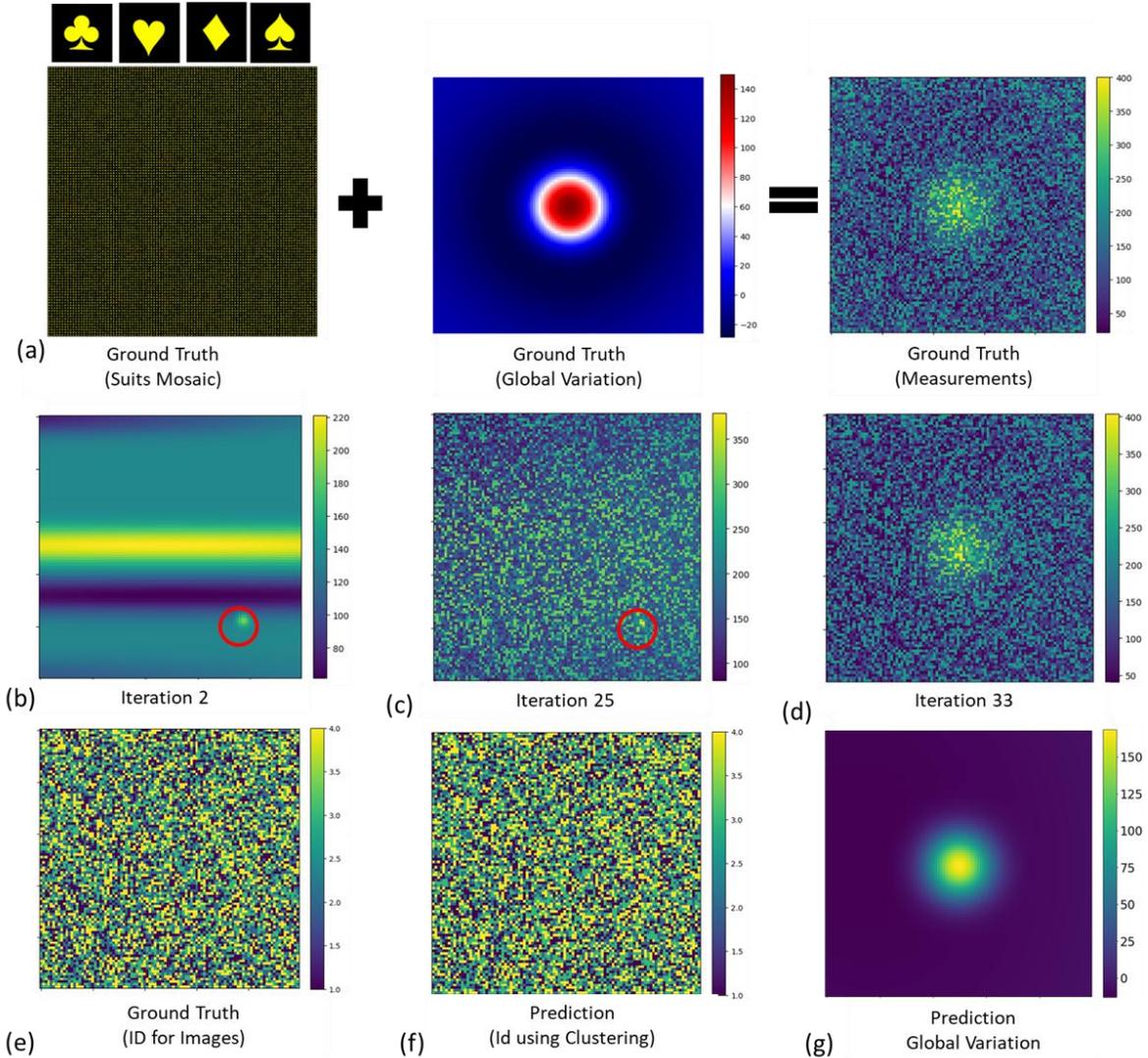

**Figure 4**: Disentangling Multi-Mechanism Complexity — Suits Dataset with Residual Stress Field

## DIVIDE on FerroSIM model

The spin–lattice FerroSim[31,32] model is a simplified framework for describing local ferroelectric behavior, representing a material as a two-dimensional lattice in which each site carries a local polarization (spin) governed by a Ginzburg–Landau–Devonshire free energy.[35] The spatial distribution of these spins defines the domain arrangement, while their evolution is described by the Landau–Khalatnikov equation, with a dynamic coefficient controlling domain wall mobility. A detailed description of the implementation of the FerroSim model and the underlying physics can be found in [31,35]. In simulations of PbTiO₃ thin films, *a*- and *c*-domains were represented by assigning site-specific defect fields and coupling parameters that determine the initial spin configuration and its resistance to switching under an external field (see Methods and Materials).

The formation of ferroelectric hysteresis loops under sinusoidal electric field switching was simulated using the FerroSim model. Hysteresis loop measurements provide access to key ferroelectric parameters, including switching work, coercive fields, and remnant polarization, all of which are reflected in the shape of loops, making this measurement central to ferroelectric characterization.[36] In this study, we selected the work of polarization switching (loop area) as the target metric. In FerroSim, a-domains act as inactive regions that cannot be switched by an external field, whereas c-domains are switchable. As a result, domain switching overwrites the initial c-domain configuration, and the influence of the starting c-domain pattern disappears after the first cycle. To capture the difference between c-domains with opposite polarization orientations, we introduced weak internal fields directed opposite to the initial polarization. These internal fields serve as proxies for imprint effects arising from defect distributions during film growth.[37] The objective of the experiment was to distinguish the contributions of a- and c-domains, acting through the imprint effect, to the hysteresis loop area.

In our simulations, the variability of local domain configurations in the FerroSim model was restricted to a single c-c domain wall separating c-domains with opposite polarization orientations, intersected by a periodic a-domain structure oriented perpendicular to this wall. The position of the *c-c* domain wall was controlled by the parameter $K$, ranging from –14 to 14 (Figure 5a), while the *a*-domain corrugation was characterized by the parameter $P$, ranging from 1 to 8, which defined the corrugation period (Figure 5d). The domain pattern dataset for the FerroSIM was formed by the overlapping (overwriting) of the c-domain pattern by the a-domain corrugation (Figure S6).

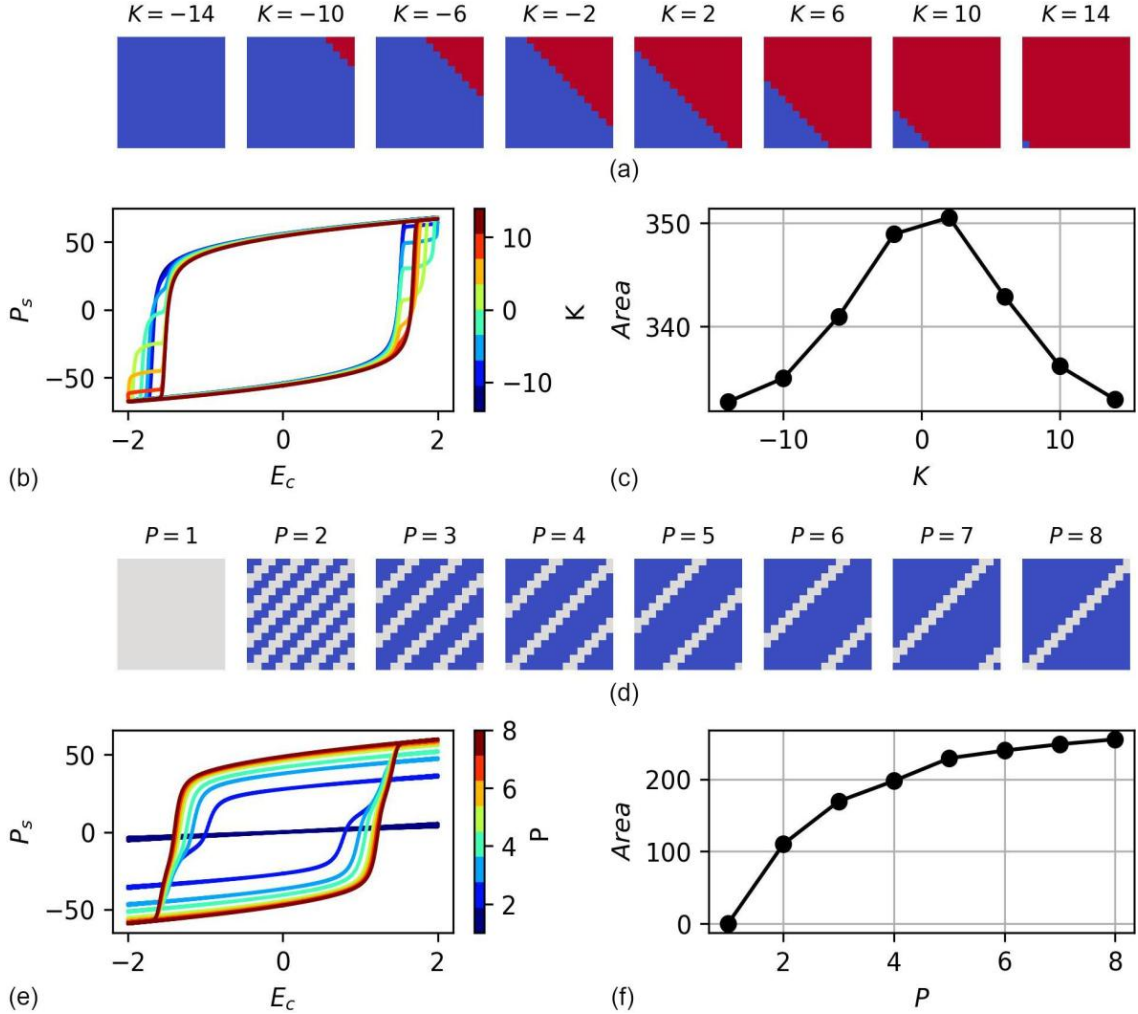

**Figure 5:** FerroSim calculations for different domain patterns used to define the ground truth. (a) c-domain arrangements for various $K$ values with corresponding (b) hysteresis loops and (c) loop areas; (d) a-domain arrangements for various $P$ values with corresponding (e) hysteresis loops and (f) loop areas

In the first step, we used the FerroSim model to calculate how the hysteresis loop shape depends on the c-domain arrangement in the absence of a-domains (Figure 5a-c). This provided the ground truth for isolating the influence of the c-domain pattern. The presence of imprint fields in domains with counter-directed polarization either suppresses or promotes local polarization switching, depending on its direction. As a result, we obtained that the loop area exhibits an extremal dependence on the position of the $c$-c domain wall, with the maximum occurring at $K = 0$, where the domain wall lies along the diagonal of the lattice (Figure 5c). To isolate the influence of the $a$-domain arrangement, we calculated the dependence of the hysteresis loop area on the parameter $P$, assuming a uniform polarization direction across all c-domains (Figure 5d-f). The results show a nonlinear increase in the hysteresis loop area with $P$, which corresponds to a simultaneous decrease in the fraction of a-domains within the lattice. When P=0, the entire lattice corresponds to non-switchable *a*-domains, and the calculated hysteresis loop in this case shows no opening, resulting in a loop area of zero.

The DIVIDE framework was employed to predict the corresponding hysteresis loop area of domain patterns, using the image patches, representing c-domain arrangement as the first channel and those representing the a-domain arrangement as the second channel (Figure S6). The model was trained on 233 domain patterns spanning the *(K, P)* exploration space. After training, the mean absolute error (MAE) between predictions (Figure 6a) and ground-truth values (Figure 6d) was $0.8 \pm 0.4$, corresponding to less than 1% of the ground-truth magnitudes for all $(P, S)$ values, except for the case of $P = 1$, where the entire lattice corresponds to a-domains, leading to the absence of hysteresis.

To analyze the efficiency of impacts separation, we plotted the dependences of loop area $(A)$ on $P$ for various fixed $K$ (constant c-domain arrangement), and conversely, the dependence of loop area on $K$ for various fixed $P$ (constant a-domain arrangement). Examples of and $A(P, K = const)$ and $A(K, P = const)$ are shown in Figure 6 b,c and Figure 6e,f correspondingly. It is evident that in both cases the shapes of the dependencies on $P$ and $K$ reproduce the ground-truth trends (Figure 5c-f), demonstrating that DIVIDE can effectively disentangle the contributions of a- and c- domains to the loop area. However, while the overall character of $A(P, K = constant)$ and $A(K, P = const)$ remain the same for different $K$, the absolute values and intersections scale with changes of fixed parameters. This indicates that the total loop area $(A_{full})$ cannot be expressed as simple addictive sum of the contributions from the a- and c- domain arrangements $(A_{full} \neq A_a + A_c)$.

Given that the shapes of $A(P, K = const)$ and $A(K, P = const)$ remain consistent across different fixed values of K and P, respectively, we conclude that while the a- and c-domain contributions are inherently independent, they may still modulate the strength of each other's influence through scaling factors:

$$A_{tot}(K, P) = \alpha(P) \cdot A_c(K) + \beta(K) \cdot A_a(P) \qquad (5)$$

where $\alpha(P)$ and $\beta(K)$ are scaling factors dependent on the opposite domain arrangement. Importantly, the *c*-domain configuration is not structurally dependent on the *a*-domain pattern, but its magnitude is scaled by the *a*-domain configuration. A true interdependence would arise only if the *c*-domain configuration itself were constrained by the *a*-domain state, i.e., if K became a function of P. The same reasoning applies in reverse: while the a-domain contribution is independent of the *c*-domain configuration, its magnitude may be scaled by it, and a true interdependence would arise only if the *a*-domain configuration were directly constrained by the *c*-domain state.

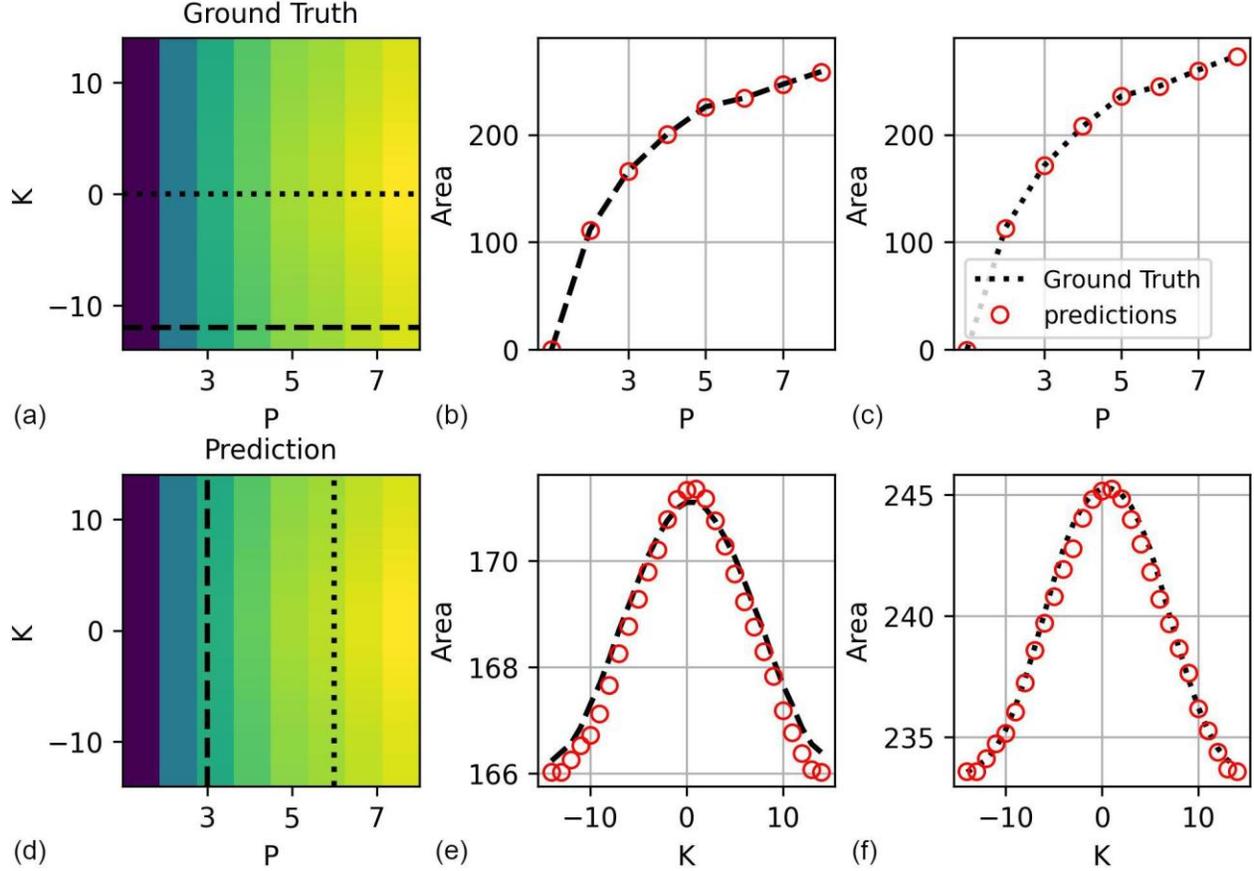

**Figure 6.** Distribution of the hysteresis loop area across the $(P, K)$ exploration space for (a) ground truth and (d) DIVIDE predictions. Panels (b, c) show the dependence of the loop area on P (encoding the a-domain arrangement) for fixed K values indicated by dashed lines in (a). Panels (e, f) show the dependence on K (encoding the c-domain arrangement) for fixed P values indicated by dashed lines in (d).

In our example, once the model is trained, we can predict the loop area for multiple realizations of *(K, P)*. Given the design of the exploration space, where the a-domain structure is superimposed on the c-domain pattern, we assume *β(K)=1*, meaning that the a-domain contribution enters additively and independently of *K*. Under this assumption, the equation 1 can be solved computationally. The "basic" contributions of $A_c(K)$ and $A_d(K)$ , together with the scaling factor $\alpha(P)$, are shown in Figure 7. In our case, the scaling coefficient exhibited an approximately linear dependence on *P* . However, this behavior should be regarded as a specific feature of the chosen domain-structure parameterization rather than a universal property

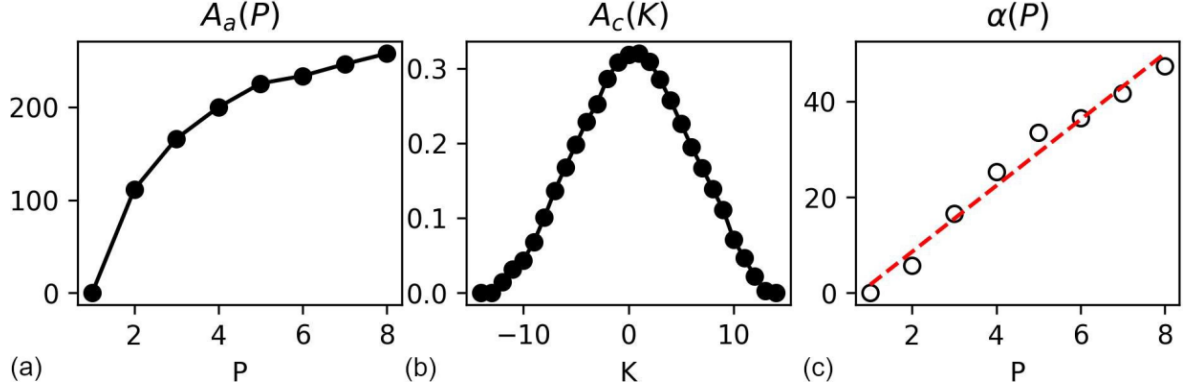

**Figure 7.** Dependencies of the basic contributions from the (a) a-domain pattern, (c) c-domain pattern, and (b) the scaling factor $\alpha(P)$ that modulates the strength of the c-domain contribution.

Summarizing the FerroSIM experiment, the DIVIDE framework demonstrated the ability to separate contributions from different domain types even in the non-additive case, where one property influences another through a scaling coefficient. This highlights the strength of the algorithm in uncovering the underlying mechanisms rather than simply fitting absolute values.

**Experimental PTO film.**

In the next step, we applied the DIVIDE framework to PFM data measured on a PTO film. The experiment was performed on the SS-PFM dataset, which is widely used for benchmarking workflows for the automated experiment and PFM data analysis.[38–40] This dataset consists of local electromechanical hysteresis loops acquired by an atomic force microscope probe. These PFM loops serve as proxies for the macroscopic ferroelectric *P-E* loops.

The area of the PFM hysteresis loops was used as the target property (Figure 8d). As input to DIVIDE, we provided two channels of 12×12px image patches, each representing the distribution of a specific structural trait around the loop measurement point (Figure 8c). The first channel encodes the a-domain arrangement as a binarized image, where 1 corresponds to an a-domain pixels and 0 to a *c-domain* pixels (Figure 8a). The second channel, the *c-domain* pattern, represents the polarization distribution within the c-domains (Figure 8b). The raw PFM signals used to construct the binarized channels are provided in the Supplementary Information (Figure S7). We employed the same objective as in the FerroSim simulations, to separate the influence of the *a-domain* arrangement from that of the polarization distribution within the *c-domains*.

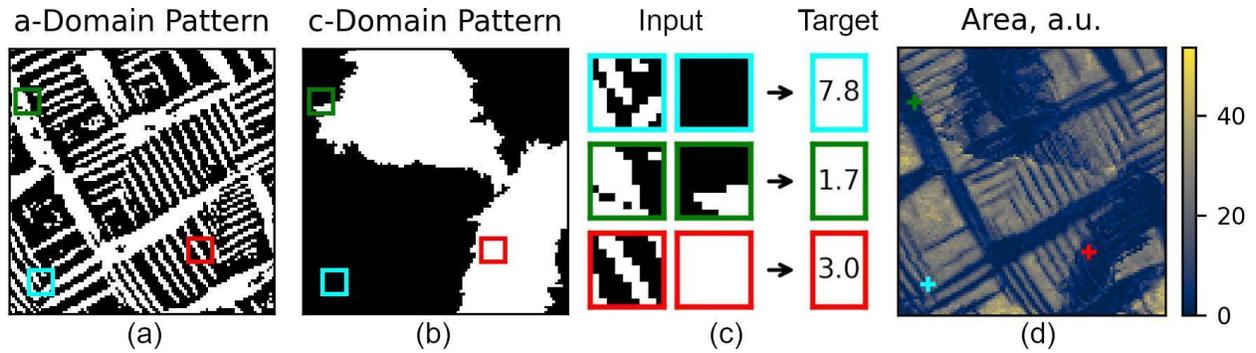

**Figure 8.** Structural images used as inputs for DIVIDE: (a) a-Domain Pattern, representing the arrangement of a-domains within the PFM scan; (b) c-Domain Pattern, showing the polarization distribution within the out-of-plane c-domains; (c) examples of input patches with corresponding scalar targets (PFM loop areas); and (d) spatial distribution of PFM loop areas across the PFM image.

The DIVIDE model with the same architecture as in the FerroSim example, was trained for 500 epochs with a learning rate of 0.01. The resulting average MAE was $2.25 \pm 1.98$, corresponding to about 18% of the average loop area across the scan. It is important to note that, due to the large contrast between values in the c- and a-domains, the relative precision compared to the absolute values can vary significantly.

Unlike the FerroSim investigation, in the case of real experimental data the domain patterns are not parameterized. To evaluate the influence of each channel, we followed an analogous approach by fixing the input in the second channel, with the fixed patch chosen based on physical considerations. In our case, to obtain the distribution of loop areas across the scan based on the a-domain arrangement, we fixed the c-domain pattern channel to represent uniform polarization in the c-domains. The resulting distributions of loop area for upward (Figure 9a) and downward (Figure 9b) polarization in the c-domains clearly correlate with the a-domain arrangements but differ significantly in their absolute values. Hysteresis loops corresponding to the case where all *c-domains* possess downward polarization have substantially smaller areas than those where all *c-domains* show upward polarization (Figure 9c). This asymmetry is most likely attributable to the imprint effect. A similar difference in loop area was observed for the distributions based on the polarization arrangement in *c-domains* "inside" and "outside" the *a-domains* (Figure 9d-f). Here, c-domains "inside" the a-domain represent subsurface c-domains located beneath the surface a-domains. The nonzero loop area experimentally measured and predicted by DIVIDE within the a-domains can thus be attributed to the contribution of these subsurface *c-domains*.

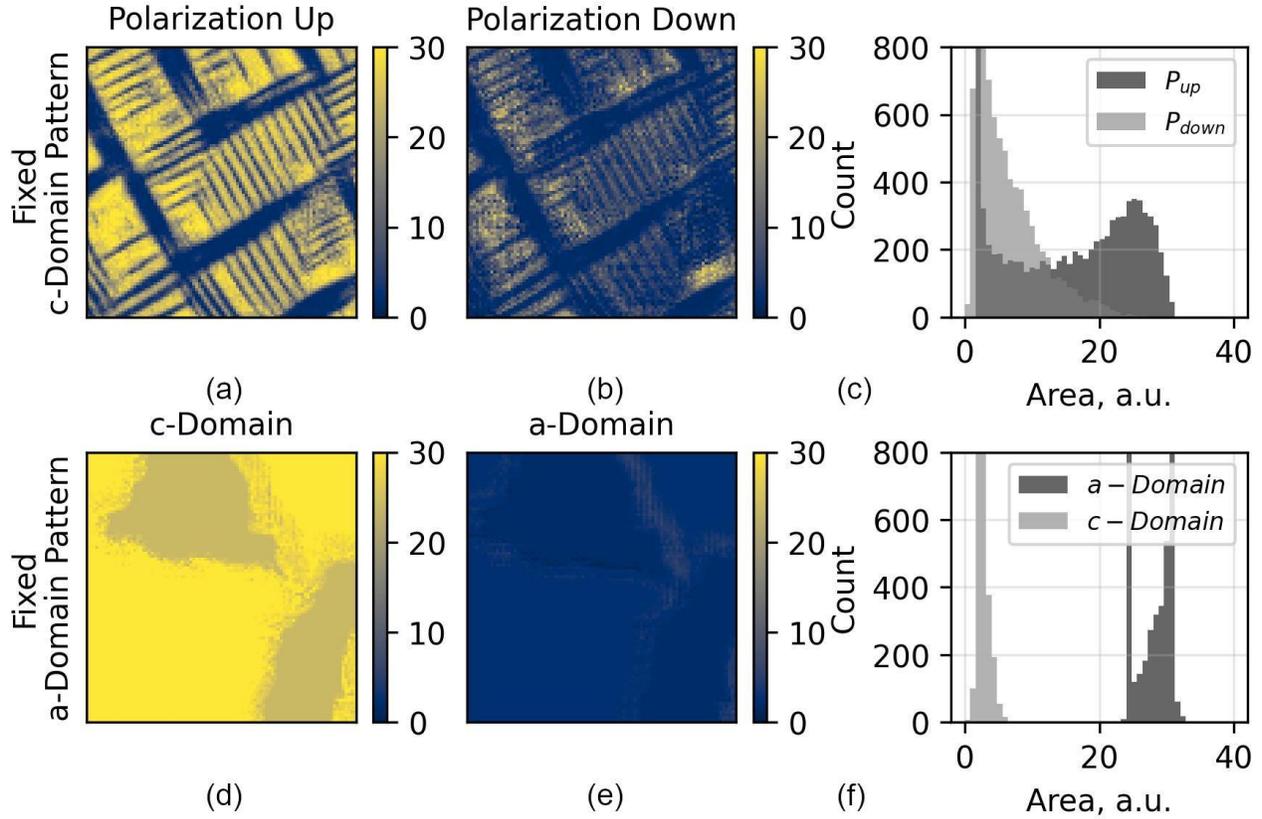

**Figure 9.** Predicted distributions of hysteresis loop areas based on (a, b) a-domain patterns and (d, e) c-domain patterns under fixed values of the other channel, with (c, f) showing the corresponding histograms.

The ability of the DIVIDE framework to separate the contributions of different mechanisms to a target property creates unique opportunities for the theoretical optimization of structural features toward desired target values. In the case of PFM, the trained DIVIDE model can not only quantify the impact of the a-domain arrangement and the polarization distribution within the c-domains but also address the reverse problem by predicting loop areas for artificial domain patterns, thereby guiding the optimization of domain structures. As a demonstration, we used the trained DIVIDE model to predict the distribution of loop areas for a real a-domain pattern combined with artificial c-domain arrangements (Figure 10). The results show that the trained model can capture the interplay between different domain types and accurately predicting the resulting loop-areas distributions, opening broad opportunities for domain engineering.

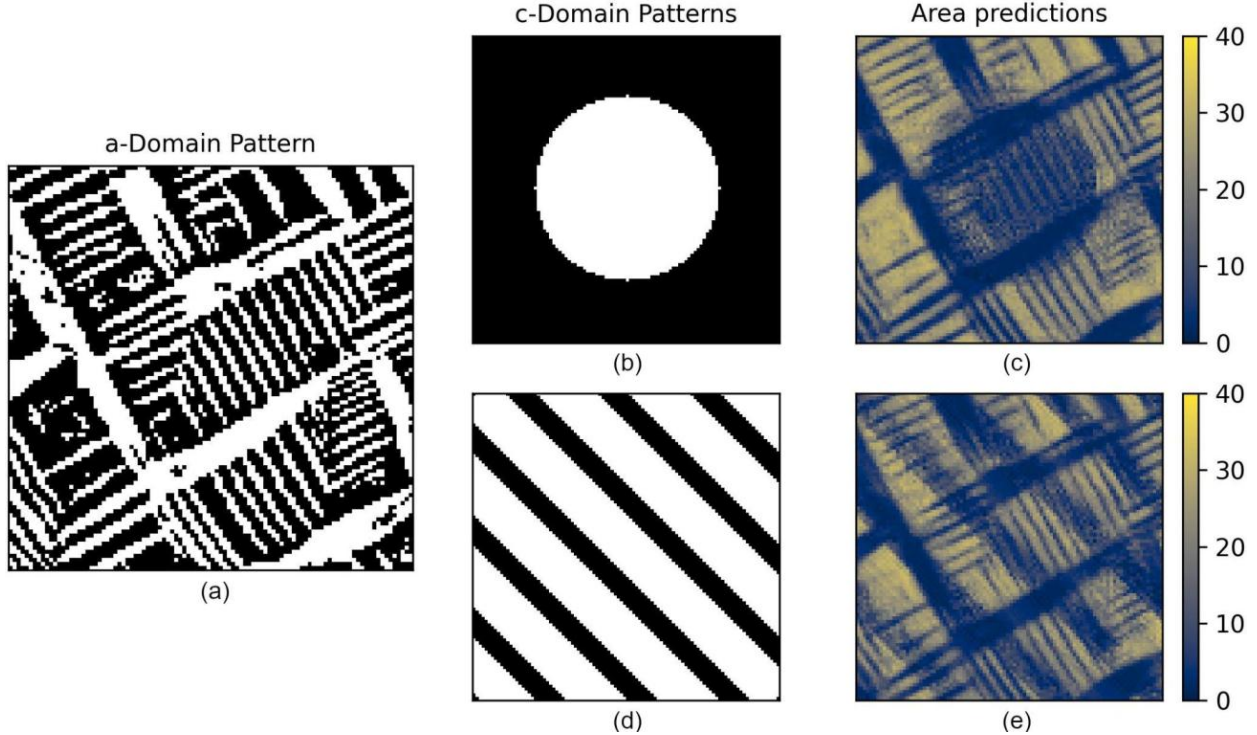

**Figure 10.** (c,e) Predicted distributions of PFM hysteresis loop areas across domain patterns encoded by (a) the real a-domain distribution and (b,d) artificial c-domain patterns

## METHODS

### Modeling Framework

We model the scalar output $y \in \mathbb{R}$ as governed by n independent mechanisms. For each mechanism $k$, a dedicated encoder $f_k(x; \theta_k)$ produces a latent vector $z_k(x) \in \mathbb{R}^{\{d_k\}}$. These latent representations are concatenated to form a joint vector $z(x) = [z_1(x), z_2(x), \ldots, z_n(x)] \in \mathbb{R}^D$, which serves as the input to a Gaussian Process (GP) with a kernel $K(z, z')$. The GP captures dependencies in the output space while respecting the modular latent structure. Each encoder is implemented as either a convolutional neural network (CNN) or a multilayer perceptron (MLP), depending on the modality of the input. The GP component uses a radial basis function (RBF) kernel. To ensure identifiability, we normalize each latent component to have zero mean and unit variance across the dataset. Additionally, we impose $n-1$ anchor constraints, such as fixing $z_1(x_1), = 0$, to resolve translational ambiguity in the latent space. In cases where low-dimensional auxiliary data (e.g., spatial coordinates or composition) is available, we normalize and append this information to the latent representation. This hybrid latent vector combines learned features with physically interpretable context, enabling the GP to capture both data-driven patterns and physically meaningful variation, which is especially useful when context variables encode known structure such as spatial locality or boundary effects. The model is trained end-to-end by maximizing the variational evidence lower bound (ELBO). All models are implemented in PyTorch, and GP regression is handled via the GPyTorch library, using variational sparse GPs with inducing points.

**Implementation Details and Error Characterization**

In our implementation, convolutional neural networks (CNNs) are consistently employed as feature extractors for all patch-based input modalities. Depending on the task, the model may use one or multiple modalities, such as symbolic suit textures or RGB image patches. Each input is processed by a dedicated CNN comprising two convolutional layers with ReLU activation, followed by adaptive average pooling and a linear projection to a 16-dimensional latent space. The resulting latent vectors are concatenated and augmented with normalized spatial coordinates in the range $[-1, 1]$, forming the input to a variational Gaussian Process.

The GP employs a Matern-5/2 kernel with automatic relevance determination (ARD) and a constant mean function, optionally initialized from prior knowledge of suit-composition combinations. We adopt a variational approximation with 50 inducing points, jointly learned during training. The model is trained end-to-end using the Adam optimizer (learning rate = 0.01) over 500 iterations with a batch size of 256. The training objective is the negative variational evidence lower bound (ELBO):

$$L_{ELBO} = E_{q(f)}[\log p(y \mid f)] - KL[q(f) \parallel p(f)] \qquad (6)$$

where $q(f)$ is the variational posterior, $p(f)$ is the GP prior, and $p(y \mid f)$ is the Gaussian likelihood. Output targets are standardized (zero mean, unit variance) during training and re-scaled for evaluation. Unless stated otherwise, all predictions are performed over the entire $100 \times 100$ spatial grid (10,000 total points). For training, a fixed subset of at most 100 points is used, either selected randomly or incrementally through active learning. In the standard setting, this subset is randomly sampled at the beginning and held fixed during training. In the active learning setting, training begins with as few as 2 datapoints, and new samples are added iteratively based on an acquisition policy (e.g., uncertainty-based or spatially penalized selection). No separate test dataset is used; rather, generalization is evaluated directly over the full spatial grid at each iteration.

Primary sources of modeling error include finite CNN capacity, limited inducing point resolution, and approximation in the variational posterior. These are mitigated through joint optimization of all components and uncertainty-guided adaptive sampling.

**Benchmark Problem Setup**

We evaluate the proposed framework on two synthetic Both benchmarks consist of a scalar output formed from the additive combination of a spatial field and a local image-based mechanism, with only the total output observed. The model receives both spatial coordinates and image patches as inputs and is tasked with learning disentangled, interpretable representations. In the first benchmark setting, the scalar target is constructed as a combination of a deterministic linear spatial gradient and a categorical image-based contribution (Figure 2a) derived from uniform color patches (red, green, or blue), each associated with a constant base value (150, 110, and 50, respectively). In the second benchmark, we move from a simple RGB patch-based dataset to a more structured and semantically rich image-based dataset composed of playing card suits (spades, hearts, diamonds, and clubs). This suits-based dataset has previously been used as a benchmark problem for evaluating deep kernel learning (DKL) models, due to its combination of categorical variation and spatial structure. We also increase the complexity of the underlying spatial variation. Instead of a simple linear variation in the x and y directions, we now embed a physically inspired, non-linear spatial field that mimics residual stress behavior typically observed in welded materials. The synthetic field is created by superimposing two Gaussian profiles: a narrow, high-magnitude tensile peak and a wider, low-magnitude compressive trough, both centered at the same spatial location. Mathematically, the residual stress field is defined as:

$$\sigma(x,y) = \sigma_{tensile} * \exp\left(-\frac{(x-x_0)^2 + (y-y_0)^2}{2r_{tensile}^2}\right) + \sigma_{compressive} \tag{7}$$
$$* \exp\left(-\frac{(x-x_0)^2 + (y-y_0)^2}{2r_{compressive}^2}\right)$$

where:

- $(x_0, y_0)$ is the weld center,
- $\sigma_{tensile} = 200$ MPa, $r_{tensile}=10$,
- $\sigma_{compressive} = -50$ MPa, $r_{compressive} = 30$.

The generated residual stress field is aligned with the Suits based grid, with a one-to-one mapping between each color patch and its corresponding residual stress value. The final training label at each location is formed by combining the suits-based mechanism with the spatial stress variation.

**FerroSIM and PFM model setup**

In the FerroSIM spin–lattice model, ferroelectric materials are represented as a two-dimensional lattice in which each site carries a local polarization.[35] The local domain arrangement in the FerroSIM model was encoded on a 15×15 lattice, with each site assigned either to an a-domain or a c-domain. Local polarizations were oriented in the *zy* plane, where the *z*-axis represents the out-of-plane polarization direction, while the *y*-axis the in-plane component. The *a*-domains were modeled as regions with a strong defect field aligned along the *y*-axis, set to be 30 times higher than the coercive field ($E_c$). This approach effectively locks the polarization vectors within *a*-domains in-plane and prevents their switching under an out-of-plane external field. A weak defect field was introduced inside the c-domains, equal to $0.1E_c$ and oriented opposite to the polarization, to model the imprint effect[37,41,42] The depolarization coefficient ($\alpha_{dep}$) was set to 0.2. A sinusoidal voltage waveform of two periods with an amplitude of $2E_c$ was applied to model polarization switching.

The thin PTO film fabricated by chemical vapor deposition was used as a model ferroelectric system[43] The pre-acquired dataset consists of PFM hysteresis loops measured on a 100×100 grid over a 3×3 μm² area using band-excitation switching spectroscopy PFM (SS-PFM) on a Cypher scanning probe microscope (Oxford Instruments).

**CONCLUSION**

We have presented DIVIDE, a framework for learning from data generated by multiple independent mechanisms, each influencing a shared scalar output. By combining modular deep encoders with a structured Gaussian Process, DIVIDE preserves the independence of contributing factors and enables interpretable prediction and representation. The design supports the incorporation of structured priors, such as spatial trends, and integrates low-dimensional physical inputs like coordinates directly into the latent space.

Through experiments on synthetic datasets with known mechanisms ranging from simple categorical signals to complex, multi-scale spatial fields, we have demonstrated DIVIDE's ability to recover and separate these influences with high accuracy. The use of active learning further improves data efficiency, allowing the model to identify mechanisms with minimal supervision. Across all cases, DIVIDE maintains clear separation in latent space, avoids entanglement, and supports stable convergence under structured GP modeling.

Beyond synthetic benchmarks, we applied DIVIDE to ferroelectric systems using both the spin-lattice FerroSim model and experimental PFM data from $PbTiO_3$ films. In FerroSim, DIVIDE successfully disentangled the non-additive contributions of a- and c-domains to hysteresis loop area. Applied to experimental PFM loops, DIVIDE further demonstrated that it can separate the effects of surface a-domain arrangement from subsurface c-domain polarization upon the area, while also reproducing imprint-related asymmetries. Importantly, the trained model was able to predict loop-area distributions for artificial domain patterns, illustrating its potential for domain-structure optimization and design.

Taken together, these results suggest that DIVIDE is a powerful and versatile framework for scientific and engineering domains where measurements result from overlapping but independent mechanisms. Its modular, interpretable architecture and compatibility with domain knowledge make it a strong candidate for extending machine learning into settings where understanding the source of variation is as important as predicting its outcome.

**Acknowledgements**

The authors would like to acknowledge Rama K. Vasudevan for access to the FerroSim ferroelectric simulation framework, and Yongtao Liu for providing the PFM dataset used in this work. This research was primarily supported by National Science Foundation Materials Research Science and Engineering Center (MRSEC) program through the UT Knoxville Center for Advanced Materials and Manufacturing (DMR-2309083).

# Supplemental Information

## S1. Benchmark Design and Ground Truth Construction

To evaluate disentanglement under controlled conditions, we constructed two synthetic benchmark problems combining two spatial field types and two local image modalities. Each benchmark generates scalar-valued outputs through an additive composition of a spatial mechanism $f_{spatial}(x, y)$, and a local mechanism $f_{local}(x, y)$. Only the total output is observed, given by:

$$z(x, y) = f_{spatial}(x, y) + f_{local}(x, y) \qquad \text{-(1)}$$

The model is provided with normalized spatial coordinates $(x, y)$ and a corresponding image patch at each location, and it is tasked with recovering the individual components and predicting the scalar output.

*Benchmark 1: Linear Spatial Field + Uniform Color Patches*

A deterministic linear gradient spans the spatial domain with output values from 10 to 300. Each image patch is uniformly colored (red, green, or blue), adding a fixed constant offset. This benchmark tests basic additive disentanglement.

### 1.1.1 Benchmark 2: Gaussian Residual Stress Field + Symbolic Image Patches

A nonlinear residual stress field, formed by the superposition of Gaussian peaks and valleys, defines the spatial mechanism. This is combined with symbolic image patches (e.g., hearts, spades) with random rotation and shear. The task challenges the model's ability to generalize, disentangle, and quantify uncertainty across modalities. The synthetic field is created by superimposing two Gaussian profiles: a narrow, high-magnitude tensile peak and a wider, low-magnitude compressive trough, both centered at the same spatial location. Mathematically, the residual stress field is defined as:

$$\sigma(x, y) = \sigma_{tensile} * \exp\left(-\frac{(x-x_0)^2 + (y-y_0)^2}{2r_{tensile}^2}\right) + \sigma_{compressive} * \exp\left(-\frac{(x-x_0)^2 + (y-y_0)^2}{2r_{compressive}^2}\right)$$

where:

- $(x_0, y_0)$ is the weld center,
- $\sigma_{tensile} = 200$ MPa, $r_{tensile} = 10$,
- $\sigma_{compressive} = -50$ MPa, $r_{compressive} = 30$.

## S2. Model Architectures

This architecture consists of a CNN encoder for the input image patch, which produces a 16-dimensional latent vector. Normalized spatial coordinates (x, y) are directly appended to this latent representation. The combined vector serves as input to a Gaussian Process, which models the scalar output.

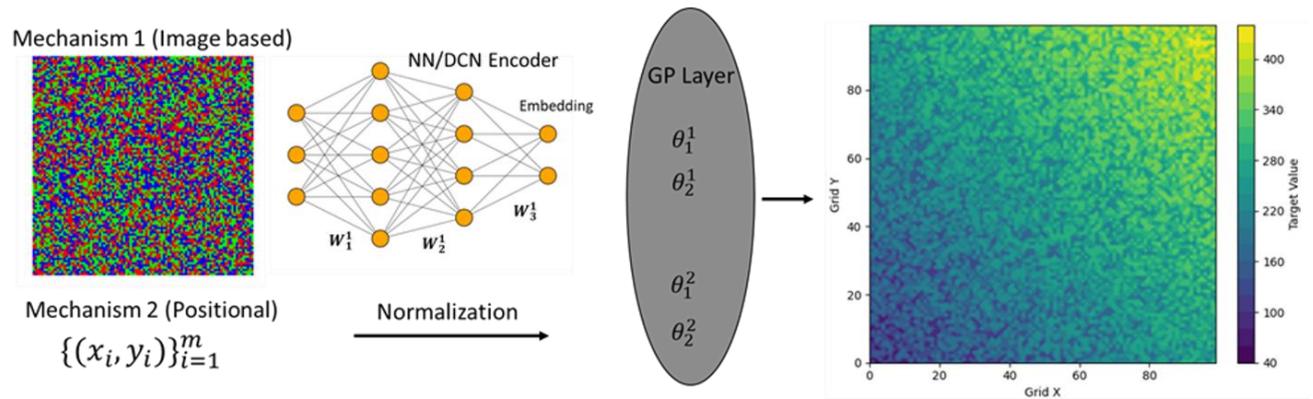

**Figure S1:** Two-Mechanism Model (Benchmark 1)

The three-mechanism model uses two independent CNN branches: one for RGB patches and another for symbolic suit patches. These latent features are concatenated with spatial coordinates to form a unified latent representation. A structured GP with a physically informed prior mean operates over this space to predict the scalar output.

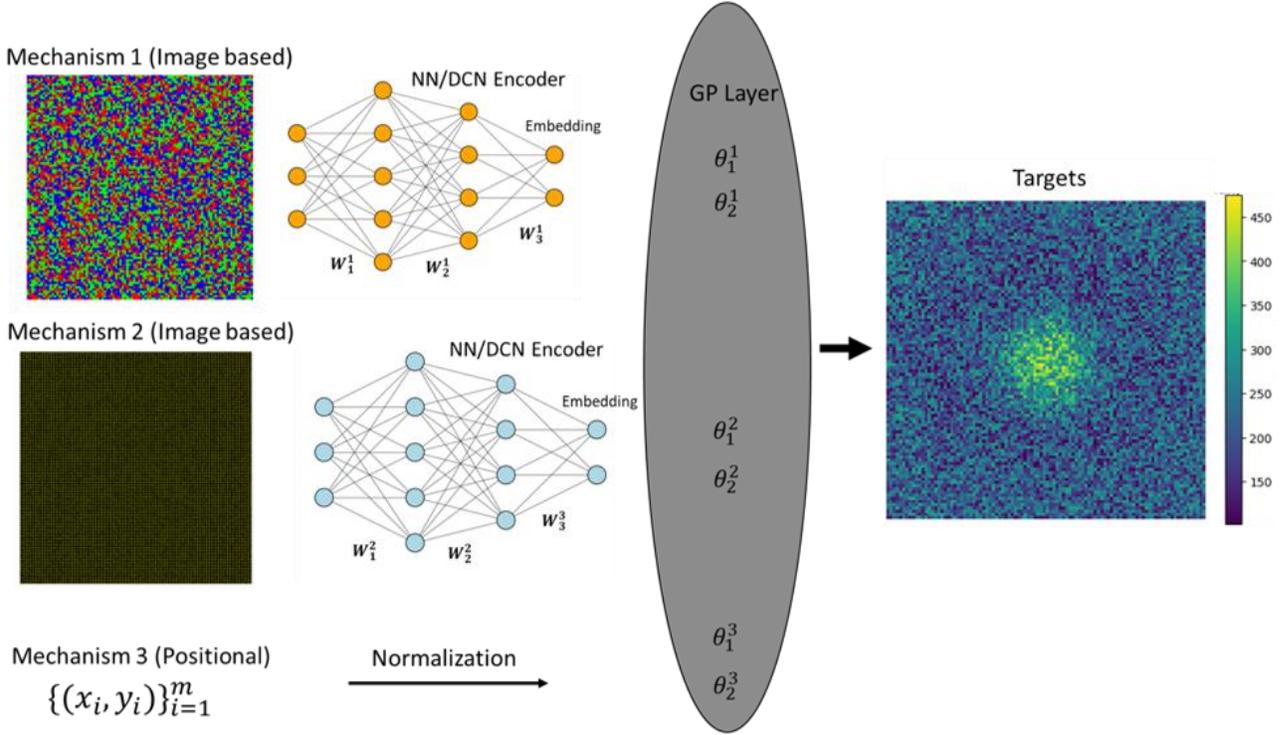

**Figure S2:** Multi-Mechanism Model (Three-Component Case)

## S3. Active Learning Implementation Details

We implemented an active learning loop to sequentially sample data points that improve model predictions with minimal supervision. The loop begins with just two randomly selected datapoints. At each iteration, the model evaluates uncertainty across the domain using Gaussian Processes conditioned on both spatial coordinates and encoded patch features.

### 1.1.2 Uncertainty-Guided Sampling

Uncertainty estimates are computed as the predictive standard deviation from the GP posterior. To avoid oversampling from the domain edges, where GP uncertainty is naturally high due to fewer neighbors, a spatial penalty is added to suppress acquisition scores at boundary points. This encourages the selection of informative locations from the interior regions and accelerates convergence.

### 1.1.3 Learning Dynamics

Initially, the model captures rough global trends via spatial coordinates and some local variation from the image patches. However, with very limited data, the GP tends to favor smooth interpolants dominated by spatial structure. After approximately 10 acquisition steps, the model shifts to recognize local effects more accurately, reconciling both sources of variation and reducing predictive uncertainty significantly.

## S4. Robustness analysis for Benchmark 1

To further evaluate the robustness of our active learning framework, we investigated two additional scenarios. In the first case, we removed all spatial variation by setting the x and y contributions to zero, isolating the effect of patch-dependent behavior. In the second case, we introduced significant measurement noise by adding zero-mean Gaussian noise with a standard deviation of 50 to a linear spatial variation. Despite these challenging conditions, the model consistently converged to the ground truth. For the no spatial variation case, the patch-driven mean values were accurately recovered within just 5 iterations, highlighting the model's ability to identify local features efficiently. In the high-noise scenario, although the overall trend appeared flattened due to the added noise, the model still managed to capture the underlying behavior closely after 15 iterations. These results confirm that the proposed active learning scheme remains effective even under reduced spatial signal or high-noise conditions, reinforcing its applicability in realistic, noisy environments. This analysis is presented in Figure S3: (a) corresponds to the no x, y variation case, while panel (b) shows the results under high measurement noise. At least for the simple case of Benchmark 1, the overall performance was highly satisfactory. The model converged to the correct solution efficiently, even under challenging conditions. These results demonstrate that the framework is robust, with active learning enabling convergence in just a few iterations. Moreover, noise in the observations was handled gracefully, indicating that the method is well-suited for practical applications where such imperfections are common.

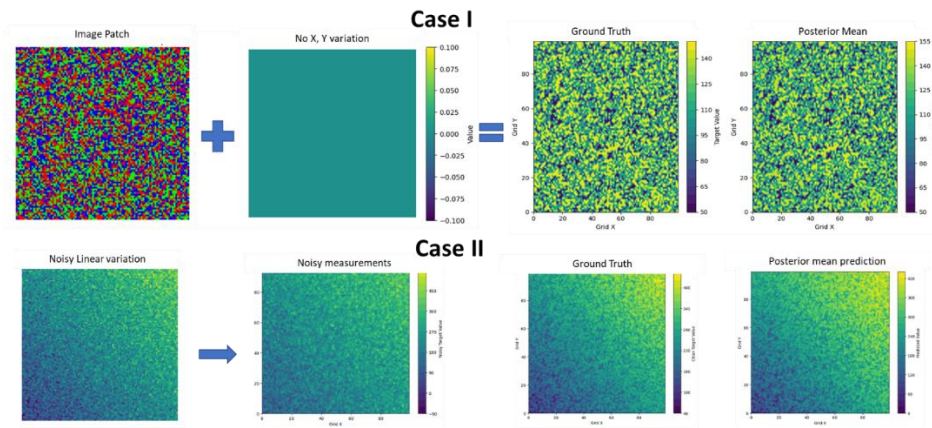

**Figure S3**: Robustness analysis under two challenging conditions. (a) No spatial variation: the model quickly recovers patch-driven outputs within 5 iterations. (b) High-noise scenario: despite added Gaussian noise ($\sigma = 50$), the underlying spatial trend is capture

### S5. Benchmark 1.1 RGB mosaic with residual stress

We further increase the complexity of the underlying spatial variation from Benchmark 1. Each patch in the mosaic is a uniform color block. Instead of a simple linear variation in the x and y directions, we now embed a physically inspired, non-linear spatial field that mimics residual stress behavior typically observed in welded materials.

We repeat the active learning procedure with this dataset. As before, the categorical component (suit identity) is quickly recovered, passing the patch-level sanity check. However, the spatial component poses a significant challenge. Even after 100 active learning iterations, the model fails to accurately learn the underlying residual stress field. We observe frequent convergence to local optima in the standard GP framework, particularly due to the multi-scale nature of the signal. The second row of Figure S4 visually highlights this failure mode.

To address the failure of standard GPs in capturing the complex spatial variation, we introduce a Structured Gaussian Process (sGP) that incorporates domain knowledge directly into the model through a custom prior mean function. Based on physical intuition from residual stress behavior in welded materials, we assume the presence of a Gaussian-like tensile peak in the spatial domain. This is encoded by defining a structured mean function of the form:

$$\mu(x, y) = A * \exp\left(-\frac{(x - x_c)^2 + (y - y_c)^2}{2\omega^2}\right)$$

where $(x_c, y_c)$ is the center of the peak, $A$ is the amplitude, and $\omega$ is the width (standard deviation). These parameters are treated as learnable and adapt during training. Importantly, this structured mean is applied only to the spatial components (i.e., the last two dimensions corresponding to $x$ and $y$ of the input), while the color-based mechanism is modeled with a separate constant mean.

Using this setup, we find that the model converges to a solution in just 15 active learning iterations. The recovered spatial variation accurately captures the central tensile peak. However, the broader and weaker compressive region remains under-represented, suggesting that while structured priors significantly improve learning efficiency and stability, capturing multi-scale effects still presents challenges. This result underscores the importance of including physically informed priors when learning from complex spatial phenomena.

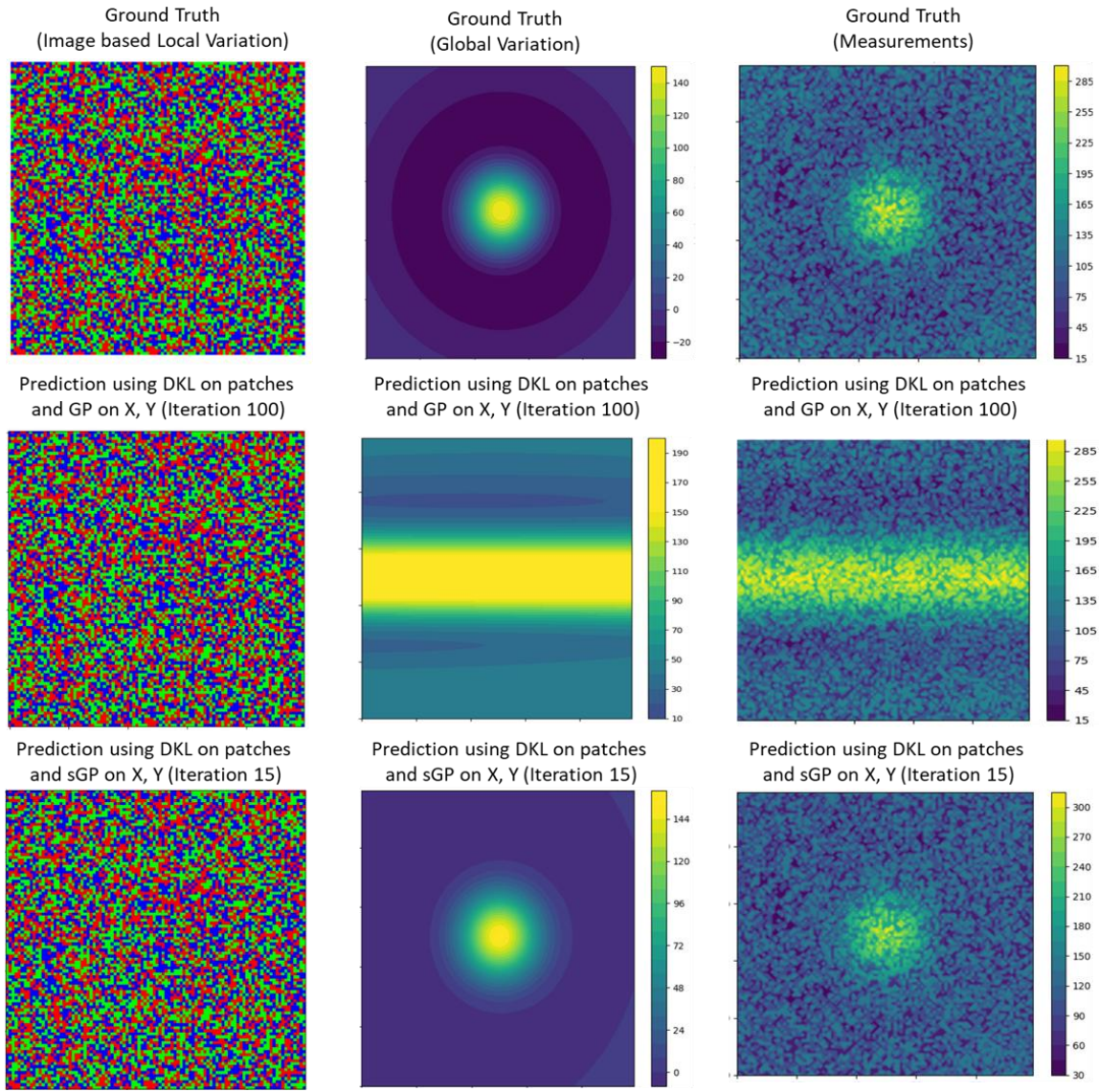

**Figure S4:** Disentangling Complex Spatial Variations Using Structured Gaussian Processes

### S6. Extension to Three Mechanisms

We construct a dataset comprising three distinct mechanisms: (1) a simple RGB color-based image mosaic, (2) a suit-based image patch dataset, and (3) a hidden spatial field with nonlinear variation modeled as a combination of Gaussian peaks. This setup introduces two categorical, image-based mechanisms and one continuous, spatial mechanism. To handle this complexity, we employ a multi-branch architecture, where two separate CNNs independently encode the RGB and symbolic patches. These latent representations are concatenated with normalized spatial coordinates and passed to a structured Gaussian Process (sGP) with a domain-informed mean. The detailed architecture is in Section S2.

Figure S5 presents the learning results for the three-mechanism case. Despite the compounded difficulty, the model is able to disentangle all three underlying mechanisms effectively. The suit and RGB-based patch classifications are correctly recovered, and the learned latent space accurately reflects their categorical identities.

Two representative trials are shown. In the second trial, the model successfully converges to a solution that captures both the center and the overall shape of the positive Gaussian peak in the residual stress field, achieving similar performance to the two-mechanism case described earlier. However, in the first trial, while the model is able to localize the center of the Gaussian peak, it fails to capture the broader compressive region of the field. This is consistent with previous observations that the model tends to underrepresent low-magnitude, distributed features in the presence of sharp, high-magnitude signals.

In both trials, training was conducted for a maximum of 100 active learning iterations. Importantly, the image-based components, RGB and suit-based, were consistently and correctly identified in both cases, confirming the stability and reliability of the proposed architecture when extended to more complex, multi-mechanism scenarios.

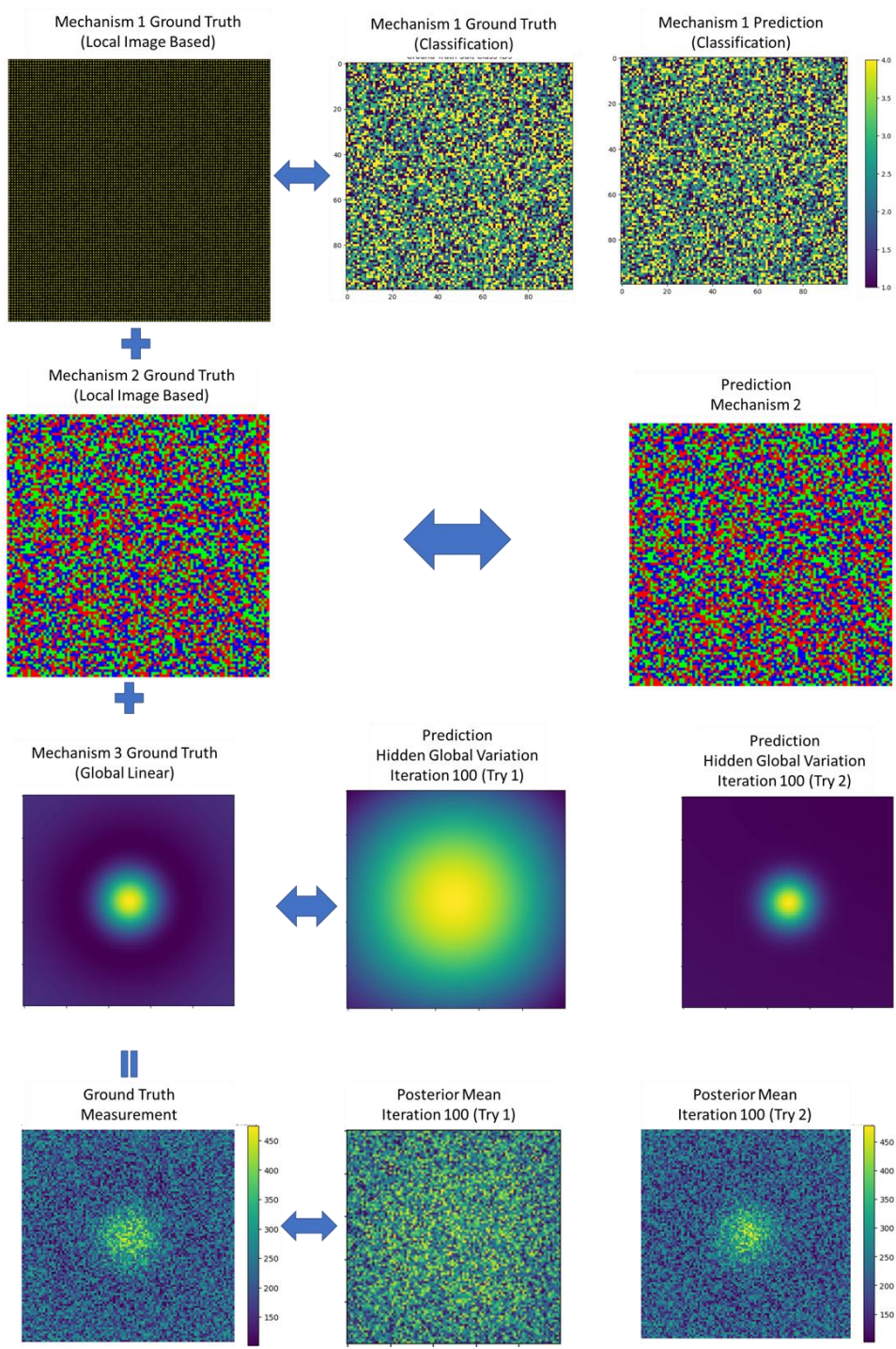

**Figure S5**: Disentanglement Results for the Three-Mechanism Case

**S6. Ferrosim Model**

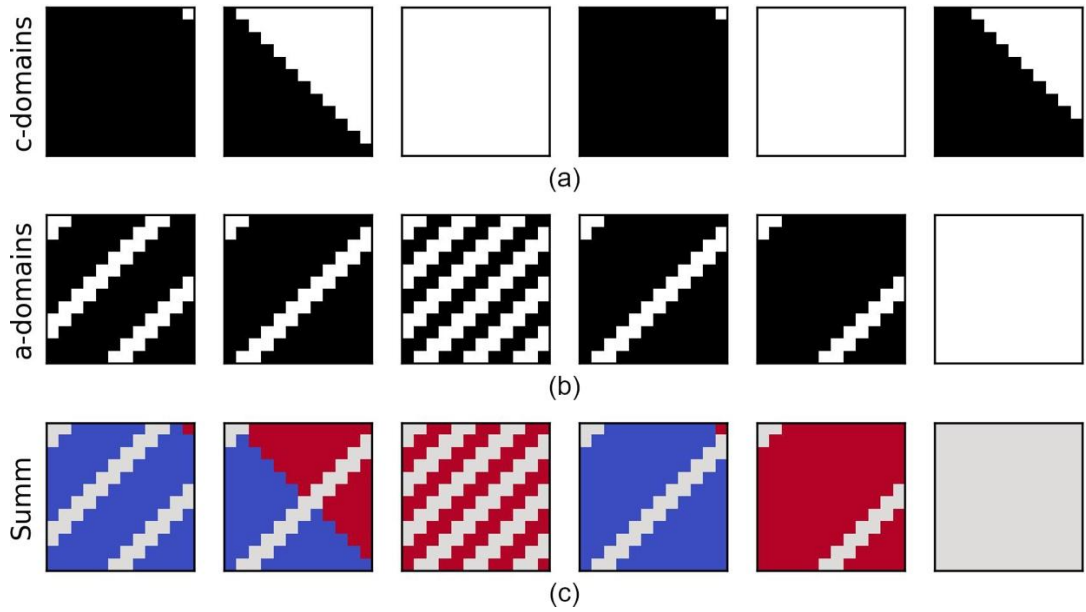

**Figure S6:** Examples of image patches representing the local arrangement of (a) c-domains, (b) a-domains, and (c) the combined domain patterns obtained by overlapping them, which were used for the FerroSim calculations. Image patches (a) and (b) were employed as inputs for the DIVIDE model.

## 1.2 S7. PTO film

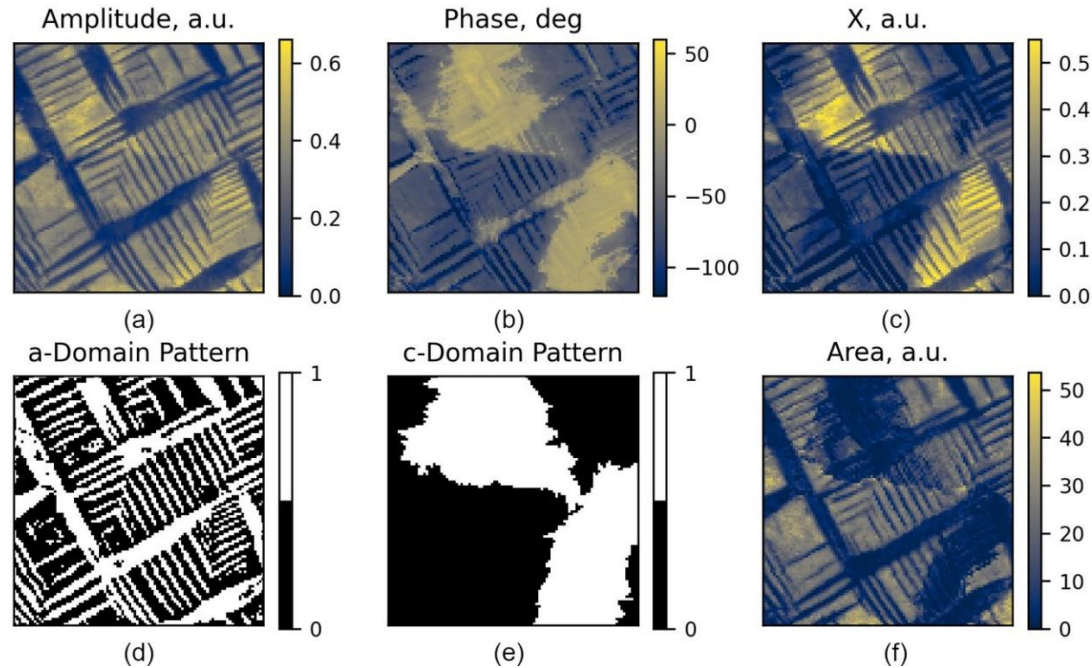

**Figure S7:** Local domain arrangement in a PTO film measured by PFM: (a) amplitude, (b) phase, (c) $X = amp \times cos(phase)$, and (f) loop area of the PFM hysteresis loops measured across the scan. Panels (d) and (e) show the binarized a-domain arrangement and the polarization distribution in c-domains, respectively, extracted from the raw signals using standard morphological transformations.